\documentclass[a4paper,twoside]{article}

\usepackage{epsfig}
\usepackage{subcaption}
\usepackage{calc}
\usepackage{booktabs}           % professional-quality tables
\usepackage{multirow}           % tabular cells spanning multiple rows
\usepackage{dblfloatfix}
\usepackage{amssymb}
\usepackage{amstext}
\usepackage{amsmath}
\usepackage{amsthm}
\usepackage{multicol}
\usepackage{pslatex}
\usepackage{apalike}
\usepackage[bottom]{footmisc}
\usepackage{SCITEPRESS}     % Please add other packages that you may need BEFORE the SCITEPRESS.sty package.

\begin{document}

\title{Evaluation of Induced Expert Knowledge in Causal Structure Learning by NOTEARS}

% \author{\authorname{Jawad Chowdhury\sup{1}\orcidAuthor{0000-0000-0000-0000}, Rezaur Rashid\sup{2}\orcidAuthor{0000-0000-0000-0000} and Gabriel Terejanu\sup{3}\orcidAuthor{0000-0000-0000-0000}}
% \affiliation{\sup{1}College of Computing and Informatics, University of North Carolina at Charlotte, North Carolina, USA}
% \affiliation{\sup{2}Department of Computing, University of North Carolina at Charlotte, North Carolina, USA}
% \email{\{mchowdh5, mrashid1, gabriel.terejanu\}@uncc.edu}
% }
%
\author{\authorname{Jawad Chowdhury, Rezaur Rashid and Gabriel Terejanu}
\affiliation{Dept. of Computer Science, University of North Carolina at Charlotte, Charlotte, NC, USA}
\email{mchowdh5@uncc.edu, mrashid1@uncc.edu, gabriel.terejanu@uncc.edu}
}


\abstract{Causal modeling provides us with powerful counterfactual reasoning and interventional mechanism to generate predictions and reason under various what-if scenarios. However, causal discovery using observation data remains a nontrivial task due to unobserved confounding factors, finite sampling, and changes in the data distribution. These can lead to spurious cause-effect relationships. To mitigate these challenges in practice, researchers augment causal learning with known causal relations. The goal of the paper is to study the impact of expert knowledge on causal relations in the form of additional constraints used in the formulation of the nonparametric NOTEARS. We provide a comprehensive set of comparative analyses of biasing the model using different types of knowledge. We found that (i) knowledge that correct the mistakes of the NOTEARS model can lead to statistically significant improvements, (ii) constraints on active edges have a larger positive impact on causal discovery than inactive edges, and surprisingly, (iii) the induced knowledge does not correct on average more incorrect active and/or inactive edges than expected. We also demonstrate the behavior of the model and the effectiveness of domain knowledge on a real-world dataset.}

\onecolumn \maketitle \normalsize \setcounter{footnote}{0} \vfill

\section{\uppercase{Introduction}}

Machine learning models have been breaking records in terms of achieving higher predictive accuracy. Nevertheless, out-of-distribution (OOD) generalization remains a challenge. One solution is adopting causal structures \cite{lake2017building} to constrain the models and remove spurious correlations. The underlying causal knowledge of the problem of interest can significantly help with domain adaptability and OOD generalization \cite{magliacane2017domain}. Furthermore, causal models go beyond the capability of correlation-based models to produce predictions. They provide us with the powerful counterfactual
reasoning and interventional mechanism to reason under various what-if
scenarios \cite{pearl2009causality}.

Two of the most prominent approaches in observational causal discovery are constraint-based and score-based methods \cite{spirtes2000causation,pearl1995theory,colombo2012learning,chickering2002optimal,ramsey2017million}.
Although these methods are quite robust if the underlying assumptions are true, they are computationally expensive and their computational complexity increases with the number of system variables due to the combinatorial nature of the DAG constraint. NOTEARS \cite{zheng2018dags} tackles this problem with an algebraic characterization of acyclicity which reduces the combinatorial problem to a continuous constrained optimization. Different approaches \cite{yu2019dag,lachapelle2019gradient,ng2019masked,zheng2020learning} have been proposed as the nonlinear or nonparametric extensions of this linear continuous optimization, which provides flexibility in modeling different causal mechanisms.
%

% %
Learning the causal structure purely based on observational data is not a trivial task due to various limitations such as finite sampling, unobserved confounding factors, selection bias, and measurement errors \cite{cooper1995causal,elkan2001foundations,zadrozny2004learning}. These can result in spurious cause-effect relationships. To mitigate these challenges in practice, researchers augment causal learning with prior causal relations as featured in software packages such as CausalNex\footnote{https://github.com/quantumblacklabs/causalnex}, 
causal-learn\footnote{https://https://github.com/cmu-phil/causal-learn}, 
% CausalNex~\shortcite{causalnex2021toolkit} 
% and DoWhy\footnote{https://microsoft.github.io/dowhy/}. 
bnlearn~\cite{scutari2009learning},
gCastle~\cite{zhang2021gcastle},
and DoWhy~\cite{sharma2020dowhy}.
Heindorf et al.~\cite{heindorf2020causenet} in their work attempts to construct the first large scale open domain causality graph that can be included in the existing knowledge bases. The work further analyze and demonstrates the benefits of large scale causality graph in causal reasoning. Given a partial ancestral graph (PAG), representing the qualitative knowledge of the causal structure, Jaber et al.~\cite{jaber2018causal} in their study compute the interventional distribution from observational data. Combining expert knowledge with structural learning further constrains the search space minimizing the number of spurious mechanisms \cite{wei2020dags} and researchers often leverage these background knowledge by exploiting them as additional constraints 
for knowledge-enhanced event causality identification \cite{liu2021knowledge}. O’Donnell et al.~\cite{o2006causal} use expert knowledge as prior probabilities in learning Bayesian Network (BN) and Gencoglu and Gruber~\cite{gencoglu2020causal} use the linear NOTEARS model to incorporate knowledge to detect how different characteristics of the COVID-19 pandemic are causally related to each other. 
Different experts' causal judgments can be aggregated into collective ones \cite{bradley2014aggregating} and Alrajeh et al.~\cite{alrajeh2020combining} in their work, studied how these judgments can be combined to determine effective interventions. An interesting exploration by Andrews et al.~\cite{andrews2020completeness} defines tiered background knowledge and shows that with this type of background knowledge the FCI algorithm~\cite{spirtes2000causation} is sound and complete. 

However, understanding how to effectively incorporate and evaluating the impact of induced knowledge is yet to be explored and we believe knowledge regarding this can mitigate some of the challenges of observational causal discovery. Human expertise can play a vital role to assess the learned model in causal structure learning \cite{bhattacharjya2021cause,li2021guided}. In practice, human assessment and validation process often take place in an iterative or sequential manner \cite{holzinger2016interactive,xin2018accelerating,yang2019study}. In structure learning, this is more realistic for a sufficiently large causal network where one can learn, validate, and induce newly formed knowledge-set in the learning process following sequential feedback loops. The goal of this paper is not to create a new causal discovery algorithm but rather to study this iterative interaction between prior causal knowledge from domain experts that takes the form of model constraints and a state-of-the-art causal structure learning algorithm. Wei et al.~\cite{wei2020dags} have been the first to augment NOTEARS with additional optimization constraints to satisfy the Karush-Kuhn-Tucker (KKT) optimality conditions and
Fang et al.~\cite{fang2020low} in their work leverages the low rank assumption in the context of causal DAG learning by augmented NOTEARS that shows significant improvements.
However, none of them have studied the impact of induced knowledge on causal structure learning by augmenting NOTEARS with the optimization constraints. For completeness, in Section~\ref{section:methodology}, we do provide our formulation of nonparametric NOTEARS \cite{zheng2020learning} with functionality to incorporate causal knowledge in the form of known direct causal and non-causal relations. Nevertheless, in this work, we aim to study the impact of expert causal knowledge on causal structure learning.

The main contributions are summarized as follows. (1) We demonstrate an iterative modeling framework to learn causal relations, impose causal knowledge to constrain the causal graphs, and further evaluate the model's behavior and performance. (2) We empirically evaluate and demonstrate that: (a) knowledge that corrects model's mistake can lead to statistically significant improvements, (b) constraints on active edges have a larger positive impact on causal discovery than inactive edges, and (c) the induced knowledge does not correct on average more incorrect active and/or inactive edges than expected. 
Finally, we illustrate the impact of additional knowledge in causal discovery on a real-world dataset.

This paper is structured as follows: Section~\ref{section:background} introduces the background on causal graphical models (CGMs), score-based structure recovery methods, and a study using the score-based approach formulated as a continuous optimization and its recent nonparametric extension. In Section~\ref{section:methodology}, we present our extension of the nonparametric continuous optimization to incorporate causal knowledge in structure learning and detail the proposed knowledge induction process. Section~\ref{section:experiments} shows the empirical evaluations and comparative analyses of the impact of expert knowledge on the model's performance. Finally, in Section~\ref{section:discussion}, we summarize our findings and provide a brief discussion on future work.

\section{\uppercase{Background}} \label{section:background}

In this section, we review the basic concepts related to causal structure learning and briefly cover a recent score-based continuous causal discovery approach using structural equation models (SEMs).

\subsection{Causal Graphical Model (CGM)}
A directed acyclic graph (DAG) is a directed graph without any directed cyclic paths \cite{spirtes2000causation}. A causal graphical model CGM($P_X,\mathcal{G}$) can be defined as a pair of a graph $\mathcal{G}$ and an observational distribution $P_X$ over a set of random variables $X=(X_1,\ldots,X_d)$. The distribution $P_X$ is Markovian with respect to $\mathcal{G}$ where $\mathcal{G}=(V,E)$ is a DAG that encodes the causal structures among the random variables $X_i \in X$ \cite{peters2017elements}. The node $i \in V$ corresponds to the random variable $X_i \in X$ and edges $(i,j) \in E$ correspond to the causal relations encoded by $\mathcal{G}$. In a causal graphical model, the joint distribution $P_x$ can be factorized as $p(x) = \prod_{i=1}^d p(x_i|x_{pa_i}^\mathcal{G})$ where $X_{pa_i}^\mathcal{G}$ refers to the set of parents (direct causes) for the variable $X_i$ in DAG $\mathcal{G}$ and for each $X_j \in X_{pa_i}^\mathcal{G}$ there is an edge $(X_j \rightarrow X_i) \in E$ \cite{peters2017elements}.

\subsection{Score-based Structure Recovery}

In a structure recovery method, given $n$ i.i.d. observations in the data matrix $\mathbf{X} = [\mathbf{x_1}|\ldots|\mathbf{x_d}] \in \mathbb{R}^{n\times d}$, our goal is to learn the underlying causal relations encoded by the DAG $\mathcal{G}$. Most of the approaches follow either a constraint-based or a score-based strategy for observational causal discovery. A score-based approach typically concentrates on identifying the DAG model $\mathcal{G}$ that fits the observed set of data $\mathbf{X}$ according to some scoring criterion $S(\mathcal{G},X)$ over the discrete space of DAGs $\mathbb{D}$ where $\mathcal{G} \in \mathbb{D}$ \cite{chickering2002optimal}. The optimization problem for structure recovery in this case can be defined as follows:
\begin{equation} \label{eqn:optimization_score}
\begin{aligned}
\quad & \hspace{5mm}\min_{\mathcal{G}} \quad & S(\mathcal{G},X)\\
\quad & \textrm{subject to} \quad & \mathcal{G} \in \mathbb{D} \\
\end{aligned}
\end{equation}
The challenge with Eq.~\ref{eqn:optimization_score} is that the acyclicity constraint in the optimization is combinatorial in nature and scales exponentially with the number of nodes $d$ in the graph. This makes the optimization problem NP-hard \cite{chickering1996learning,chickering2004large}.

\subsection{NOTEARS: Continuous Optimization for Structure Learning}

NOTEARS~\cite{zheng2018dags} is a score-based structure learning approach which reformulates the combinatorial optimization problem to a continuous one through an algebraic characterization of the acyclicity constraint in Eq.~\ref{eqn:optimization_score} via trace exponential. This method encodes the graph $\mathcal{G}$ defined over the $d$ nodes to a weighted adjacency matrix $W = [w_1|\ldots|w_d] \in \mathbb{R}^{d\times d}$ where $w_{ij}\neq 0$ if there is an active edge $X_i \rightarrow X_j$ and $w_{ij}=0$ if there is not. The weighted adjacency matrix $W$ entails a linear SEM by $X_i=f_i(X) + N_i=w_i^T X + N_i$; where $N_i$ is the associated noise. The authors define a smooth score function on the weighted matrix as $h(W) = \text{tr}(e^{W \circ W}) - \text{d}$ where $\circ$ is the Hadamard product and $e^M$ is the matrix exponential of M. This embedding of the graph $\mathcal{G}$ and the characterization of acyclicity turns the optimization in Eq.~\ref{eqn:optimization_score} into its equivalent:
\begin{equation} \label{eqn:optimization_notears}
\begin{aligned}
\quad & \hspace{5mm}\min_{W \in \mathbb{R}^{d\times d}} \quad & L(W)\\
\quad & \textrm{subject to} \quad & h(W)=0 \\
\end{aligned}
\end{equation}
where $L(W)$ is the least square loss over $W$ and $h(W)$ score defines the DAG-ness of the graph.

\subsection{Nonparametric Extension of NOTEARS}

A nonparametric extension of the continuous optimization suggested by a subsequent study \cite{zheng2020learning} uses partial derivatives for asserting the dependency of $f_j$ on the random variables. The authors define $f_j \in H^1(\mathbb{R}^d) \subset L^2(\mathbb{R}^d)$ over the Sobolev space of square integrable functions whose derivatives are also square integrable. The authors show that $f_j$ can be independent of random variable $X_i$ if and only if $||\partial_i f_j||_{L^2}=0$ where $\partial_i$ denotes partial derivative with respect to the $i$-th variable. This redefines the weighted adjacency matrix with $W(f)=W(f_1,\ldots,f_d) \in \mathbb{R}^{d\times d}$ where each $W_{ij}$ encodes the partial dependency of $f_j$ on variable $X_i$. As a result, we can equivalently write Eq.~\ref{eqn:optimization_notears} as follows:
\begin{equation} \label{eqn:optimization_nonparametric}
\begin{aligned}
\quad & \hspace{5mm}\min_{f:f_j \in H^1(\mathbb{R}^{d}),\forall j \in [d]} \quad & L(f)\\
\quad & \textrm{subject to} \quad & h(W(f))=0 \\
\end{aligned}
\end{equation}
for all $X_j \in X$. Two of the general instances proposed by \cite{zheng2020learning} are: NOTEARS-MLP and NOTEARS-Sob. A multilayer perceptron having $h$ number of hidden layers and $\sigma:\mathbb{R}\rightarrow \mathbb{R}$ activation function can be defined as $M(\mathbf{X};L)=\sigma(L^{(h)}\sigma(\ldots \sigma(L^{(1)}X))$ where $L^{(l)}$ denotes the parameters associated with $l$-th hidden layer. The authors in \cite{zheng2020learning} show that if $||i\text{-th column of } L_j^{(1)}||_2=0$ then $M_j(\mathbf{X};L)$ will be independent of variable $X_i$ which replaces the association of partial derivatives in Eq.~\ref{eqn:optimization_nonparametric} and redefines the adjacency matrix as $W(\theta)$ with $W(\theta)_{ij}=||i\text{-th column of } L_j^{(1)}||_2$ where $\theta=(\theta_1,\ldots,\theta_d)$; $\theta_k$ denoting the set of parameters for the $M_k(\mathbf{X};L)$ ($k$-th MLP). With the usage of neural networks and the augmented Lagrangian method \cite{bertsekas1997nonlinear} NOTEARS-MLP solves the constrained problem in Eq.~\ref{eqn:optimization_nonparametric} as follows:
\begin{equation} \label{eqn:optimization_mlp}
\begin{aligned}
\quad & \hspace{5mm}\min_{\theta} F(\theta) + \lambda{||\theta||}_1 \\
\quad & F(\theta)= L(\theta) + \dfrac{\rho}{2} {|h(W(\theta))|}^2 + \alpha h(W(\theta)) \\
\end{aligned}
\end{equation}

\section{\uppercase{Knowledge Induction}} 
\label{section:methodology}

In our formulation, we use the multilayer perceptrons of NOTEARS-MLP proposed by \cite{zheng2020learning} as our estimators. We extend this framework to incorporate causal knowledge by characterizing the extra information as additional constraints in the optimization in Eq.~\ref{eqn:optimization_nonparametric}. 
\paragraph{Knowledge Type.} We distinguish between these two types of knowledge: (i) \emph{known inactive} is knowledge from the true inactive edges (absence of direct causal relation), and (ii) \emph{known active} is knowledge from the true active edges (presence of direct causal relation). 
\paragraph{Knowledge Induction Process.} We adopt an interactive induction process, where the expert knowledge is informed by the outcome of the causal discovery model. Namely, the knowledge is induced to correct the mistakes of the model in the causal structure, in the hope that the new structure is closer to the true causal graph. This process is applied sequentially by correcting the mistakes of the model at each step.

In the following subsections we present the formulation of the NOTEARS optimization with constrains and detail the sequential induction process.
\begin{figure}
% \begin{figure}[tb]
  \centering
  \includegraphics[width=0.98\linewidth]{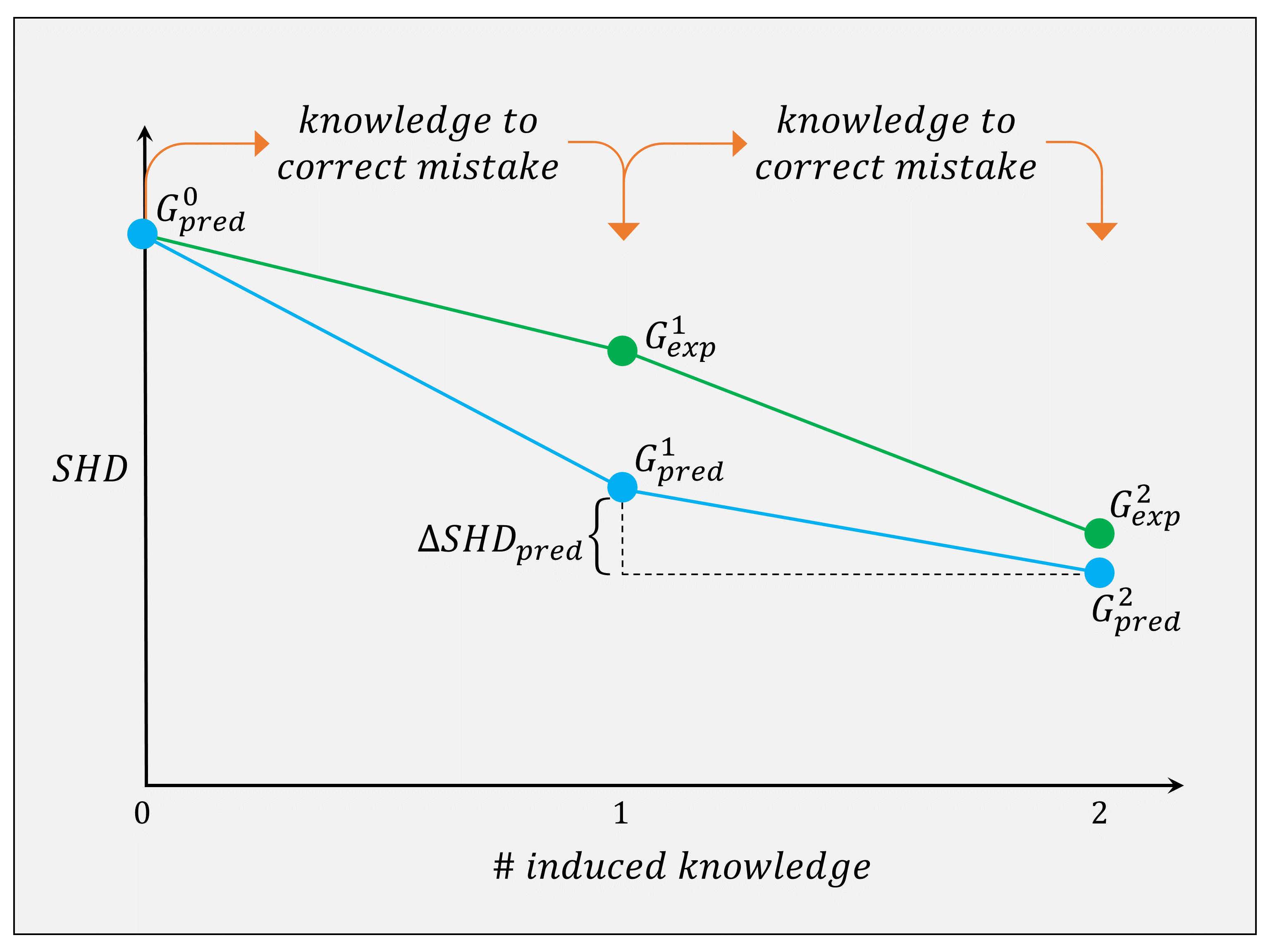}
  \caption{Knowledge induction process. We induce knowledge by carrying over the existing knowledge set along with a new random correction informed by model mistakes.}
%   \Description{}
  \label{fig:4_conditional_vs_unconditional}
\end{figure} 
\begin{figure}
% \begin{figure}[tb]
  \centering
  \includegraphics[width=0.98\linewidth]{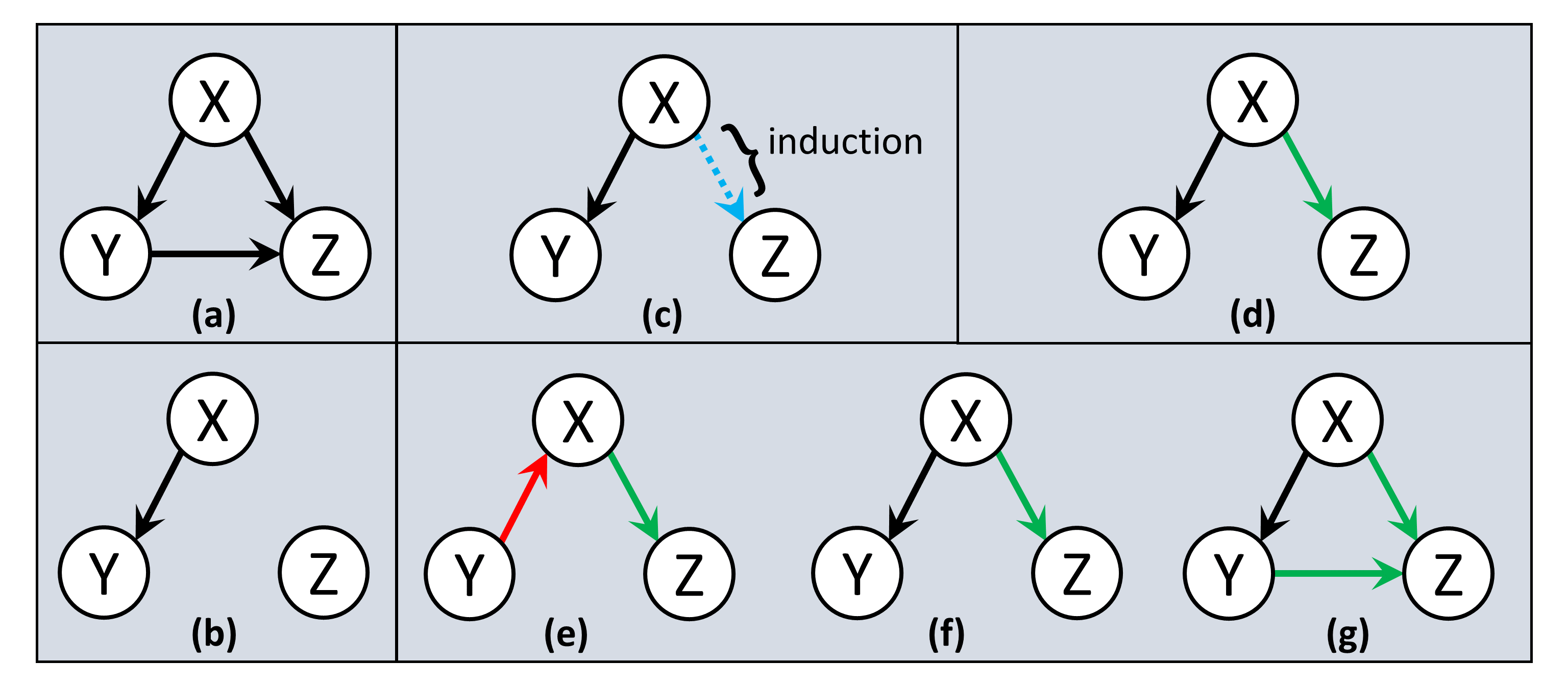}
  \caption{Expected graph formulation: (a) true graph, $\mathcal{G}_{true}$, (b) predicted graph by model at step $k$, $\mathcal{G}_{pred}^k$, (c) induced knowledge at step $(k+1)$, (d) expected graph at step $(k+1)$, $\mathcal{G}_{exp}^{k+1}$. Three different examples of many possible predicted graphs at step $(k+1)$, $\mathcal{G}_{pred}^{k+1}$ where the model performs (e) less than expectation, (f) par with expectation, and (g) more than expectation.}
%   \Description{}  
  \label{fig:4_delta_calc}
\end{figure} 
\subsection{Expert Knowledge as Constraints}
An induced knowledge associated with a true active edge, $X_i \rightarrow X_j$ (\textit{known active}) enforces the corresponding cell in the adjacency matrix to be non-zero, $[W(\theta)]_{ij}\neq 0$. We consider this knowledge as inequality constraint in our extension of the optimization such that the following statement holds:
\begin{equation} \label{eqn:constraints_ineq}
\begin{aligned}
\quad & h^p_{ineq}(W(\theta)) > 0 \\
\end{aligned}
\end{equation}
where $p$ enumerates over all the inequality constraints due to induction from the set of \textit{known active} and $h_{ineq}$ is the penalty score associated with the violation of these inequality constraints. 
On the other hand, knowledge associated with true inactive edge, $X_i \nrightarrow X_j$ (\textit{known inactive}) enforces the related cell in $W(\theta)$ to be equal to zero, $[W(\theta)]_{ij}=0$ if the induction implies there should not be an edge from $X_i$ to $X_j$. We consider this knowledge as equality constraint in our optimization such as:
\begin{equation} \label{eqn:constraints_eq}
\begin{aligned}
\quad & h^q_{eq}(W(\theta))=0 \\
\end{aligned}
\end{equation}
where $q$ enumerates over all the equality constraints, induced from the set of \textit{known inactive} and $h_{eq}$ is the penalty score associated with the violation of these equality constraints. With these additional constraints in Eqs.~\ref{eqn:constraints_ineq},~\ref{eqn:constraints_eq} we extend Eq.~\ref{eqn:optimization_nonparametric} to incorporate causal knowledge in the optimization as follows:
\begin{equation} \label{eqn:optimization_nonparametric_2}
\begin{aligned}
\quad & \hspace{5mm}\min_{f:f_j \in H^1(\mathbb{R}^{d}),\forall j \in [d]} \quad & L(f)\\
\quad & \textrm{subject to} \quad & h(W(\theta))=0, \\
\quad & \quad & h^q_{eq}(W(\theta))=0, \\
\quad & \quad & h^p_{ineq}(W(\theta)) > 0 \\
\end{aligned}
\end{equation}
NOTEARS uses a thresholding step on the estimated edge weights to reduce false discoveries by pruning all the edges with weights falling below a certain threshold. Because of this, in practice, even the equality constraints in Eq.~\ref{eqn:constraints_eq} become inequalities to allow for small weights. Finally, slack variables are introduced in the implementation to transform the inequality constraints into equality constraints (see detailed formulation in Appendix A).

By using the similar strategy suggested by Zheng et al.~\cite{zheng2020learning} with augmented Lagrangian method the reframed constrained optimization of Eq.~\ref{eqn:optimization_mlp} takes the following form:
\begin{equation} \label{eqn:optimization_mlp_2}
\begin{aligned}
\quad & \hspace{5mm}\min_{\theta} F(\theta) + \lambda{||\theta||}_1 \\
\quad & F(\theta)= L(\theta) + \dfrac{\rho}{2} {|h(W(\theta))|}^2 + \alpha h(W(\theta)) \\
\quad & + \sum_p (  \dfrac{\rho_{ineq}}{2}{|h^p_{ineq}(W(\theta))|}^2 + \alpha_p h^p_{ineq}(W(\theta))) \\
\quad & + \sum_q (  \dfrac{\rho_{eq}}{2}{|h^q_{eq}(W(\theta))|}^2 + \alpha_q h^q_{eq}(W(\theta))) \\
\end{aligned}
\end{equation}

\subsection{Sequential Knowledge Induction} 
In case of knowledge induction, the optimization is run in a sequential manner where the constraints are informed by the causal mistakes made by the model in the previous step. We start with our baseline model without imposing any additional knowledge from the true DAG and get the predicted causal graph denoted by $\mathcal{G}_{pred}^{0}$ in Figure~\ref{fig:4_conditional_vs_unconditional}. Then at each iterative step $(k+1)$, based on the mistakes in the causal graph $\mathcal{G}_{pred}^{k}$ predicted by the NOTEARS-MLP, we select one additional random piece of knowledge to correct one of the mistakes, and add it to the set of constraints identified in the previous $k$ steps, and rerun NOTEARS. We note that a batch of corrections can also be selected, however for this work we have focused on estimating the contribution of each piece of knowledge in the form of known active/inactive edge. Our observations are illustrated in Section~\ref{subsection:experiments_corrects_mistake},  Section~\ref{subsection:experiments_eq_vs_ineq}, Section~\ref{subsection:experiments_imp_vs_exp}, and Section~\ref{subsection:experiments_real_data}.
\paragraph{Expected Causal Graph.} We consider the expected causal graph, $\mathcal{G}_{exp}^{k+1}$ at step $(k+1)$ by considering the case where all the knowledge has successfully been induced without impacting any other edges. Figure~\ref{fig:4_delta_calc}d illustrates an example of how we formulate our expected graph for a particular step in the iterative process. We note that the correction might yield a directed graph (Expected Causal Graph) that is not necessary a DAG. The objective is to compare the performance between the causal graph predicted by NOTEARS and the expected causal graph. Our intuition is that the induced knowledge will probably correct additional incorrect edges, see Figure~\ref{fig:4_delta_calc}g, yielding a performance better than expected.

\section{\uppercase{Experiments}}  
\label{section:experiments}
\begin{table}
% \begin{table}[tb]
    \centering
    \caption{Performance metrics considered with their corresponding desirability.} \centering
    % \begin{center}
    \begin{tabular}{|c|c|}
    % \begin{tabular}{@{}rr@{}}
        \hline
        Metric & Desirability \\ 
        \hline
        $\Delta$FDR            & Lower is better       \\ 
        \hline
        $\Delta$TPR            & Higher is better      \\ 
        \hline
        $\Delta$FPR            & Lower is better       \\ 
        \hline
        $\Delta$SHD            & Lower is better       \\ 
        \hline
    \end{tabular}
    % \end{center}
    \label{tab:4_metric_desirability}
\end{table}

\begin{table}
% \begin{table}[tb]
    \centering
    \caption{Results for inducing redundant knowledge.} \centering 
    % \begin{center}
    \begin{tabular}{|c|c|c|}
    % \begin{tabular}{@{}rrr@{}}
        \hline
        Metric & Mean ± Stderr. & Remarks      \\ 
        \hline
        $\Delta$FDR            & -0.00030 ± 0.00017      & No harm \\ 
        \hline
        $\Delta$TPR            & -0.00035 ± 0.00027      & No harm \\
        \hline
        $\Delta$FPR            & -0.00097 ± 0.00059      & No harm \\ 
        \hline
        $\Delta$SHD            & -0.00154 ± 0.00167      & No harm \\ 
        \hline
    \end{tabular}
    % \end{center}
    \label{tab:4_no_harm}
\end{table}
%

% \begin{table}
\begin{table*}[t]
    \centering
    \caption{Results for inducing knowledge that corrects model's mistake.} \centering
    % \begin{center}
    \begin{tabular}{|c|c|c|c|}
    % \begin{tabular}{@{}rrrr@{}}
        \hline
        Metric       & Knowledge & Mean ± Stderr. & Improvement \\
        \hline
        {$\Delta$FDR} & inactive           & -0.018 ± 0.002          & Significant      \\ 
        \hline
        {$\Delta$FDR} & active             & -0.008 ± 0.001          & Significant      \\
        \hline
        {$\Delta$TPR} & inactive           & -0.007 ± 0.003          & Not significant  \\
        \hline
        {$\Delta$TPR} & active             & 0.024 ± 0.003           & Significant      \\ 
        \hline
        {$\Delta$FPR} & inactive           & -0.023 ± 0.004          & Significant      \\ 
        \hline 
        {$\Delta$FPR} & active             & -0.008 ± 0.003          & Significant      \\ 
        \hline
        {$\Delta$SHD} & inactive           & -0.032 ± 0.012          & Significant      \\
        \hline
        {$\Delta$SHD} & active             & -0.071 ± 0.011          & Significant      \\ 
        \hline
    \end{tabular}
    % \end{center}
    \label{tab:4_corrects_mistake}
\end{table*}
% \end{table}
%

To empirically evaluate the impact of additional causal knowledge on causal learning and to keep our experimental setup similar to the study in Ref.~\cite{zheng2020learning}, we have used an MLP with 10 hidden units and sigmoid activation functions. In all our experimental setup, we assume the prior knowledge is correct (agrees with the true DAG). Despite the known sensitivity of the NOTEARS algorithm to data scaling, as demonstrated in previous study \cite{reisach2021beware}, we have conducted experiments using both unscaled and scaled data to ensure the robustness of our findings and we are pleased to report that our conclusions remain unchanged regardless of the scaling of the data, indicating the stability and reliability of our results. While we present the results using the unscaled data for consistency with the original implementation of NOTEARS ~\cite{zheng2020learning}, it is important to note that our conclusions hold true even when the data is scaled.
\paragraph{Simulation.} We investigate the performance of our formulation and the impact of induced knowledge by comparing the DAG estimates with the ground truths. For our simulations with synthetic data, we have considered 16 different combinations following the simulation criteria: two random graph models, Erdos-Renyi (ER) and Scale-Free (SF), number of nodes, $d = \{10,20\}$, sample size, $n = \{200,1000\}$, edge density, $s0 = \{1d,4d\}$. For each of these combinations, we have generated 10 different random graphs or true DAGs (as 10 trials for a particular combination) and corresponding data by following nonlinear data generating process with index models (similar to the study in Ref.~\cite{zheng2020learning} for which the underlying true DAGs are identifiable. The results are summarized over all these 160 random true DAGs and datasets. In our simulations, we have considered the regularization parameter, $\lambda = 0.01$. We evaluate the performance of causal learning based on the mean and the standard error of different metrics. For statistical significance analysis, we have used t-test with $\alpha = 0.05$ as the significance level.
\paragraph{Metrics.} For the comparative analysis, we consider the following performance metrics: False Discovery Rate (FDR), True Positive Rate (TPR), False Positive Rate (FPR), and Structural Hamming Distance (SHD). However, since we are evaluating the performance over all these 160 random graphs of varying sizes, we consider Structural Hamming Distance per node (SHD/d) as our SHD measure that scales with the number of nodes (FDR, TPR, and FPR scale by definition). To evaluate the impact of induced knowledge, we calculate the differences in the metrics at different steps (where we have different sizes of induced knowledge set) and referred them as $\Delta$FDR, $\Delta$TPR, $\Delta$FPR, and $\Delta$SHD, see also Table~\ref{tab:4_metric_desirability}. For example, based on our model's prediction we calculate the impact of inducing one additional piece of knowledge on the metric SHD ($\Delta \text{SHD}_{pred}$) as follows:
\begin{equation} \label{eqn:metric_calc_pred_1}
\begin{aligned}
\Delta \text{SHD}_{pred} = \text{SHD}(\mathcal{G}_{pred}^{k+1}) - \text{SHD}(\mathcal{G}_{pred}^{k})
\end{aligned}
\end{equation}
%
% \begin{table}
\begin{table*}[!b]
    \centering
    \caption{Comparison between the impact of inducing knowledge regarding inactive vs active edges.} \centering
    % \begin{center}
    \begin{tabular}{|c|c|c|c|}
    % \begin{tabular}{@{}rrrr@{}}
        \hline
        Metric & Inactive & Active & Better \\ 
        \hline
        $\Delta$FDR            & -0.019 ± 0.002                     & -0.008 ± 0.001                   & inactive                \\ \hline
        $\Delta$TPR            & -0.007 ± 0.003                     & 0.024 ± 0.003                    & active                  \\ \hline
        $\Delta$FPR            & -0.023 ± 0.004                     & -0.009 ± 0.004                   & inactive                \\ \hline
        $\Delta$SHD            & -0.033 ± 0.013                     & -0.072 ± 0.011                   & active                  \\ \hline
    \end{tabular}
    % \end{center}
    \label{tab:4_eq_vs_ineq}
% \end{table}
\end{table*}
\paragraph{Sanity Check - Redundant Knowledge Does No Harm.} As part of our sanity check, we investigate the impact of induced knowledge that matches the causal relationships successfully discovered by the NOTEARS-MLP. Therefore, in this section, we consider the set of edges that our baseline model correctly classifies as our knowledge source. Here, we do not distinguish between the edge types of our induced knowledge (\textit{known inactive \& active}) since our goal is to investigate whether having redundant knowledge as additional constraints affects model's performance or not. The results are illustrated in Table~\ref{tab:4_no_harm}. Our empirical evaluation shows that adding redundant knowledge does not deteriorate the performance of NOTEARS-MLP. Our performed statistical test reflects that the results after inducing the knowledge from the correctly classified edge set are not statistically different than the results from the model without these knowledge inductions. However, we have noticed that the performance gets worse with highly regularized models. This is consistent with observations by Ng et al.~\cite{ng2020role} where sparse DAGs result in missing some of the true active edges.

\subsection{Knowledge that Corrects Model's Mistake} 
\label{subsection:experiments_corrects_mistake}

We first investigate the role of randomly chosen knowledge that corrects model's mistake based on the cause-effect relations of the true graph. Therefore, in this case, we consider the set of misclassified edges from the estimated causal graph as the knowledge source for biasing the model. The results are illustrated in Table~\ref{tab:4_corrects_mistake}.  
Our empirical result shows statistically significant improvements whenever the induced knowledge corrects misclassified edges in the estimated causal graph except for the case of $\Delta$TPR with \textit{known inactive} edges. However, this behavior is not totally unexpected since knowledge from \textit{known inactive} edges helps to get rid of false discoveries or false positives, which hardly have impact on true positives.  
%

% \begin{table}
\begin{table*}[t]
    \centering
    \caption{Comparison between the empirical performance vs expectation.} \centering
    % \begin{center}
    \begin{tabular}{|c|c|c|c|c|}
    % \begin{tabular}{@{}rrrrr@{}}
        \hline
        Metric       & Knowledge & Empirical & Expected & Remarks \\ 
        \hline
        {$\Delta$FDR} & inactive           & -0.019 ± 0.002     & -0.016 ± 0.002    & No difference    \\
        \hline
        {$\Delta$FDR} & active             & -0.008 ± 0.001     & -0.006 ± 0.001    & No difference    \\ 
        \hline
        {$\Delta$TPR} & inactive           & -0.007 ± 0.003     & -0.002 ± 0.003    & No difference    \\
        \hline
        {$\Delta$TPR} & active             & 0.024 ± 0.003      & 0.022 ± 0.002     & No difference    \\ 
        \hline
        {$\Delta$FPR} & inactive           & -0.023 ± 0.004     & -0.021 ± 0.004    & No difference    \\  
        \hline
        {$\Delta$FPR} & active             & -0.009 ± 0.003     & -0.007 ± 0.003    & No difference    \\ 
        \hline
        {$\Delta$SHD} & inactive           & -0.033 ± 0.013     & -0.047 ± 0.010    & No difference    \\
        \hline
        {$\Delta$SHD} & active             & -0.072 ± 0.011     & -0.056 ± 0.010    & No difference    \\ 
        \hline
    \end{tabular}
    % \end{center}
    \label{tab:4_imp_vs_exp}
\end{table*}
% \end{table}
%

\subsection{Known Inactive vs Known Active} 
\label{subsection:experiments_eq_vs_ineq}

In this subsection, we are interested in understanding the impact of different types of induced knowledge on causal discovery to correct the mistakes in the estimated causal graph. As a result, the experimental setup is similar to Section~\ref{subsection:experiments_corrects_mistake} where we consider the misclassified edge set as the knowledge source. We consider both \textit{known inactive} and \textit{known active} types of knowledge to induce separately and analyze the differences of their impact on the performance. The results are illustrated in Table~\ref{tab:4_eq_vs_ineq}.
Based on our statistical test, we have found that inducing \textit{known inactive} is more effective when we compare the performance based on FDR and FPR as misclassification of inactive edges has more impact on these metrics. On the other hand, the results show that inducing \textit{known active} is more effective on TPR as misclassification of active edges has more impact on this metric. 
Interestingly, we have found that \textit{known active} provides a significant improvement over \textit{known inactive} in terms of SHD. This can be attributed to the fact that the induced knowledge based on the true inactive edge (\textit{known inactive}) between two random variables, i.e. from $X_i$ to $X_j$ allows for two extra degrees of freedom since it is still possible to have no edge at all or an active edge from $X_j$ to $X_i$. However, the induced knowledge based on the true active edge doesn't allow any degrees of freedom. This type of knowledge is more restraining for causal graph discovery and therefore carries more information.    

\subsection{Empirical Performance vs Expectation} 
\label{subsection:experiments_imp_vs_exp}
In this subsection, we are interested in understanding whether inducing knowledge to correct model's mistakes exceeds expected improvement. The experimental setup is similar to Section~\ref{subsection:experiments_corrects_mistake} and Section~\ref{subsection:experiments_eq_vs_ineq} where we consider the misclassified edge set as the knowledge source. We have conducted the experiments using both \textit{known inactive} and \textit{known active} types of knowledge separately. The expected causal graph, $\mathcal{G}_{exp}$ is formulated in a similar manner described in Fig.~\ref{fig:4_delta_calc}. Table~\ref{tab:4_imp_vs_exp} shows the summary of the performance comparison in these cases with the expected results. 
Our statistical test shows that the induced correct knowledge does not correct on average more incorrect active and/or inactive edges than expected. Therefore, using the information from induced knowledge does not have additional impact than expected in the global optimization scheme. However, this is likely due to the fact that the structure of the expected causal graph, $\mathcal{G}_{exp}$ is not well-posed.
It's worth noting that $\mathcal{G}_{exp}$ isn't necessarily a DAG since there isn't any constraining mechanism to enforce acyclicity as compared to $\mathcal{G}_{pred}$ (NOTEARS imposes hard acyclicity constaint in the continuous optimization). Although it is to be noted here that solving an acyclicity constrained optimization problem does not guarantee to return a DAG and Ng et al.~\cite{ng2022convergence} in their study illustrates on this behavior and proposes the convergence guarantee with a DAG solution.

\subsection{Real Data} 
\label{subsection:experiments_real_data}

We evaluate the implication of incorporating expert knowledge on the dataset from study in Ref.~\cite{sachs2005causal}, which is largely used in the literature of probabilistic graphical models with a consensus network accepted by the biological community. This dataset contains the expression levels of phosphorylated proteins and phospholipids in human cells under different conditions. The dataset has $d=11$ cell types along with $n=7466$ samples of expression levels. As for the ground truth of the underlying causal graph, we considered $s_0=20$ active edges as suggested by the study~\cite{sachs2005causal}. 
We have opted for $\Delta$TPR, the percentage difference of edges in agreement (higher is better), and the percentage difference of reversed edges (lower is better) as the evaluation metrics since the performance on these metrics would indicate the significance more distinctively. Similar to the synthetic data analysis, we had $10$ trials that we used to summarize our evaluation.
Our empirical result (Mean ± Stderr.) shows: $\Delta$TPR as 0.020 ± 0.004, the percentage difference of edges in agreement as 0.393 ± 0.086, and the percentage difference of reversed edges as -0.073 ± 0.030. We have found that with the help of induced knowledge the model shows statistically significant improvement by correctly identifying more active edges and by reducing the number of edges identified in the reverse direction.  
Due to the limitation of having access only to a subset of the true active edges, our analyses could not include a comparative study on \textit{known inactive} edges as in the synthetic data case. We assume the performance could have been improved by fine-tuning the model's parameters but since our main focus of this study is entirely based on the analyses regarding the impact of induced knowledge of different types and from different sources on structure learning, we kept the parameter setup similar for all consecutive steps in the knowledge induction process.

\section{\uppercase{Conclusions}}
\label{section:discussion}

We have studied the impact of expert causal knowledge on causal structure learning and provided a set of comparative analyses of biasing the model using different types of knowledge. Our findings show that knowledge that corrects model's mistakes yields significant improvements and it does no harm even in the case of redundant knowledge that results in redundant constraints. This suggest that the practitioners should consider incorporating domain knowledge whenever available. More importantly, we have found that knowledge related to active edges has a larger positive impact on causal discovery than knowledge related to inactive edges which can mostly be attributed to the difference between the number of degrees of freedom each case reduces. This finding suggest that the practitioners may want to prioritize incorporating knowledge regarding presence of an edge whenever applicable. Furthermore, our experimental analysis shows that the induced knowledge does not correct on average more incorrect active and/or inactive edges than expected. This finding is rather surprising to us, as we have expected that every constraint based on a known active/inactive edge to impact and correct more than one edge on average. 

Our work points to the importance of the human-in-the-loop in causal discovery that we would like to further explore in our future studies.
%
%an example of which could be to analyze %the impact of induced knowledge that %contradicts the true causal structure. 
%
Also, we would like to mention that in our study we adopted hard constraints to accommodate the prior knowledge since we have assumed our priors to be correct. An interesting future direction would be to accommodate the continuous optimization with functionality to allow different levels of confidence on the priors.
%

% \vfill
\section*{\uppercase{Acknowledgements}}
Research was sponsored by the Army Research Office and was accomplished under Grant Number W911NF-22-1-0035. The views and conclusions contained in this document are those of the authors and should not be interpreted as representing the official policies, either expressed or implied, of the Army Research Office or the U.S. Government. The U.S. Government is authorized to reproduce and distribute reprints for Government purposes notwithstanding any copyright notation herein.

% If any, should be placed before the references section
% without numbering. To do so please use the following command:
% \textit{$\backslash$section*\{ACKNOWLEDGEMENTS\}}

% \vfill
\bibliographystyle{apalike}
{\small
\bibliography{main}}

% \vfill
\section*{\uppercase{Appendix}}

% If any, the appendix should appear directly after the
% references without numbering, and not on a new page. To do so please use the following command:
% \textit{$\backslash$section*\{APPENDIX\}}
%
\paragraph{A. Threshold Incorporation and Slack Variables}
\label{section:appendix_1}

In Eq.~\ref{eqn:constraints_ineq}, we have seen that our inequality constraint takes the following form:
% $$ \begin{aligned}
\[
    h^p_{ineq}(W(\theta)) > 0 \\
\]
% \end{aligned}$$
%
where $p$ enumerates over each induced knowledge associated with true active edge (\textit{known active}) $X_i \rightarrow X_j$ imposing $[W(\theta)]_{ij} \neq 0$. NOTEARS uses a thresholding step that reduces false discoveries where any edge weight below the threshold value, $w_{thresh}$ in its absolute value is set to zero. Thus, for any induction from true active edges ($X_i \rightarrow X_j$) we have the following constraint:
%
% $$
% \begin{aligned}
% %\quad & |[W(\theta)]_{ij}| \geq W_{thresh}, \\
% \quad & [W(\theta)]_{ij}^2 \geq W^2_{thresh}.
% %\quad & - [W(\theta)]_{ij}^2 + W^2_{thresh} \leq 0 \\
% \end{aligned}
% $$
\[
    [W(\theta)]_{ij}^2 \geq W^2_{thresh}.
\]
We convert inequality constraints in our optimization to equality by introducing a set of slack variables $y_p$ such that:
\begin{equation} \label{eqn:transform_constraints_ineq}
\begin{aligned}
\quad & - [W(\theta)]_{ij}^2 + W^2_{thresh} + y_p = 0 \quad & \textrm{s.t.} \quad & y_p \geq 0
\end{aligned}
\end{equation}

In a similar manner, using the threshold value, $W_{thresh}$ our equality constraints (associated with \textit{known inactive} edges) take the form as:

\begin{equation} \label{eqn:transform_constraints_eq}
\begin{aligned}
\quad & [W(\theta)]_{ij}^2 - W^2_{thresh} + y_q = 0 \quad & \textrm{s.t.} \quad & y_q \geq 0 
\end{aligned}
\end{equation}

where $q$ enumerates over each induction associated with true inactive edge $X_i \nrightarrow X_j$ imposing $[W(\theta)]_{ij} = 0$.
\paragraph{B. Additional Results and Summary Statistics}
% \begin{table}
\begin{table*}[tb]
    \centering
    \caption{Full results for inducing redundant knowledge (Sanity Check).}
    % \begin{center}
    \begin{tabular}{|c|c|c|c|c|c|}
    % \begin{tabular}{@{}rrrrrr@{}}
        \hline
        Metric       & $\Delta$ & Mean ± Stderr. & p-value & t-stat & Remarks \\ 
        \hline
        $\Delta$FDR & 1              & -0.00030 ± 0.00017      & 0.076            & -1.770          & No harm          \\
        \hline
        $\Delta$FDR & 2              & -0.00060 ± 0.00021      & 0.004            & -2.850          & No harm          \\ 
        \hline
        $\Delta$TPR & 1              & -0.00035 ± 0.00027      & 0.205            & -1.260          & No harm          \\ 
        \hline
        $\Delta$TPR & 2              & -0.00036 ± 0.00029      & 0.227            & -1.210          & No harm          \\ 
        \hline
        $\Delta$FPR & 1              & -0.00097 ± 0.00059      & 0.100            & -1.630          & No harm          \\  
        $\Delta$FPR & 2              & -0.00183 ± 0.00069      & 0.008            & -2.660          & No harm          \\ 
        \hline
        $\Delta$SHD & 1              & -0.00154 ± 0.00167      & 0.356            & -0.920          & No harm          \\ 
        \hline
        $\Delta$SHD & 2              & -0.00357 ± 0.00188      & 0.050            & -1.900          & No harm          \\ 
        \hline
    \end{tabular}
    % \end{center}
    \label{tab:full_4_no_harm}
\end{table*}
% \end{table}
%

We illustrate here the detailed performance with summary statistics of induced knowledge from our empirical evaluation ($\Delta$FDR, $\Delta$TPR, $\Delta$FPR, and $\Delta$SHD for both $\Delta$=1 and $\Delta$=2). Similar to one additional knowledge ($\Delta$=1), we calculate the impact of inducing two additional piece of knowledge ($\Delta$=2) based on our model's prediction i.e. on the metric SHD ($\Delta \text{SHD}_{pred}^{2}$) as follows:
\begin{equation} \label{eqn:metric_calc_pred_2}
\begin{aligned}
\Delta \text{SHD}_{pred}^{2} = \text{SHD}(\mathcal{G}_{pred}^{k+2}) - \text{SHD}(\mathcal{G}_{pred}^{k})
\end{aligned}
\end{equation}
Table~\ref{tab:full_4_no_harm} shows the results for inducing redundant knowledge or knowledge that is correctly classified by NOTEARS-MLP.
%

% \begin{table}
\begin{table*}[!b]
    \centering
    \caption{Full results for inducing knowledge that corrects model's mistake (Section~\ref{subsection:experiments_corrects_mistake}).} 
    % \begin{center}
    \begin{tabular}{|c|c|c|c|c|c|c|}
    % \begin{tabular}{@{}rrrrrrr@{}}
        \hline
        Metric       & $\Delta$     & Knowledge & Mean ± Stderr. & p-value & t-stat & Improvement \\ 
        \hline
        $\Delta$FDR & 1 & inactive           & -0.018, 0.002           & 3.41E-14         & -7.800          & Significant      \\
        \hline
        $\Delta$FDR & 1 & active             & -0.008, 0.001           & 2.51E-08         & -5.657          & Significant      \\ 
        \hline
        $\Delta$FDR & 2 & inactive           & -0.023, 0.003           & 2.74E-15         & -8.221          & Significant      \\ 
        \hline
        $\Delta$FDR & 2 & active             & -0.011, 0.002           & 9.06E-08         & -5.448          & Significant      \\ 
        \hline
        $\Delta$TPR & 1 & inactive           & -0.007, 0.003           & 3.10E-02         & -2.164          & Not significant   \\ 
        \hline
        $\Delta$TPR & 1 & active             & 0.024, 0.003           & 8.58E-19         & 9.191           & Significant      \\ 
        \hline
        $\Delta$TPR & 2 & inactive           & -0.001, 0.003           & 8.25E-01         & -0.222          & Not significant   \\ 
        \hline
        $\Delta$TPR & 2 & active             & 0.035, 0.004            & 1.16E-19         & 9.580           & Significant      \\ 
        \hline
        $\Delta$FPR & 1 & inactive           & -0.023, 0.004           & 3.81E-08         & -5.583          & Significant      \\ 
        \hline
        $\Delta$FPR & 1 & active             & -0.008, 0.003           & 1.21E-02         & -2.517          & Significant      \\  
        \hline
        $\Delta$FPR & 2 & inactive           & -0.021, 0.003           & 1.04E-08         & -5.845          & Significant      \\ 
        \hline
        $\Delta$FPR & 2 & active             & -0.015, 0.005           & 6.73E-03         & -2.724          & Significant      \\ 
        \hline
        $\Delta$SHD & 1 & inactive           & -0.032, 0.012           & 9.74E-03         & -2.594          & Significant      \\ 
        \hline
        $\Delta$SHD & 1 & active             & -0.071, 0.011           & 1.61E-10         & -6.522          & Significant      \\ %\cline{2-7} 
        \hline
        $\Delta$SHD & 2 & inactive           & -0.082, 0.012           & 1.93E-10         & -6.533          & Significant      \\ 
        \hline
        $\Delta$SHD & 2 & active             & -0.126, 0.016           & 3.41E-14         & -7.875          & Significant      \\ 
        \hline
    \end{tabular}
    % \end{center}
    \label{tab:full_4_corrects_mistake}
\end{table*}
% \end{table}
%
Table~\ref{tab:full_4_corrects_mistake} shows the detailed results for inducing knowledge that corrects model's mistake.
%
% \begin{table}
\begin{table*}[tb]
    \centering
    \caption{Full results of comparison between the impact of inducing knowledge regarding inactive vs active edges. (Section~\ref{subsection:experiments_eq_vs_ineq}).} 
    % \begin{center}
    \begin{tabular}{|c|c|c|c|c|c|c|}
    % \begin{tabular}{@{}rrrrrrr@{}}
        \hline
        Metric       & $\Delta$ & Inactive & Active & p-value & t-stat & Better \\ 
        \hline
        $\Delta$FDR & 1          & -0.019 ± 0.002    & -0.008 ± 0.001  & 1.30E-04         & -3.85           & Inactive        \\
        \hline
        $\Delta$FDR & 2          & -0.023 ± 0.002    & -0.011 ± 0.001  & 5.58E-04         & -3.47           & Inactive        \\ 
        \hline
        $\Delta$TPR & 1          & -0.007 ± 0.003    & 0.024 ± 0.003   & 8.13E-14         & -7.57           & Active          \\
        \hline
        $\Delta$TPR & 2          & -0.001 ± 0.003    & 0.035 ± 0.004   & 2.84E-13         & -7.43           & Active          \\ 
        \hline
        $\Delta$FPR & 1          & -0.023 ± 0.004    & -0.009 ± 0.004  & 7.28E-03         & -2.69           & Inactive        \\ 
        \hline
        $\Delta$FPR & 2          & -0.021 ± 0.004    & -0.015 ± 0.005  & 3.23E-01         & -0.99           & No difference   \\ 
        \hline
        $\Delta$SHD & 1          & -0.033 ± 0.013    & -0.072 ± 0.011  & 1.90E-02         & 2.35            & Active          \\  
        \hline
        $\Delta$SHD & 2          & -0.082 ± 0.013    & -0.126 ± 0.016  & 3.28E-02         & 2.14            & Active          \\ 
        \hline
    \end{tabular}
    % \end{center}
    \label{tab:full_4_eq_vs_ineq}
\end{table*}
% \end{table}
%
Table~\ref{tab:full_4_eq_vs_ineq} shows the detailed results of the difference between the impact of \textit{`known inactive'} (knowledge induced from inactive edges) and \textit{`known active'} (knowledge induced from active edges) using misclassified edge set as the knowledge source.
%
% \begin{table}
\begin{table*}[tb]
    \centering
    \caption{Full results of comparison between the empirical performance vs expectation (Section~\ref{subsection:experiments_imp_vs_exp}).} 
    % \begin{center}
    \begin{tabular}{|c|c|c|c|c|c|c|c|}
    % \begin{tabular}{@{}rrrrrrrr@{}}
        \hline
        Metric       & $\Delta$         & Knowledge & Empirical & Expected & p-value & t-stat & Remarks \\ 
        \hline
        $\Delta$FDR & 1 & inactive           & -0.019 ± 0.002     & -0.016 ± 0.002    & 0.51             & -0.65           & No difference    \\ \hline 
        $\Delta$FDR & 1 & active             & -0.008 ± 0.001     & -0.006 ± 0.001    & 0.21             & -1.25           & No difference    \\ \hline 
        $\Delta$FDR & 2 & inactive           & -0.023 ± 0.002     & -0.025 ± 0.002    & 0.60             & 0.53            & No difference    \\ \hline
        $\Delta$FDR & 2 & active             & -0.011 ± 0.002     & -0.010 ± 0.002    & 0.75             & -0.32           & No difference    \\ \hline
        $\Delta$TPR & 1 & inactive           & -0.007 ± 0.003     & -0.002 ± 0.003    & 0.22             & -1.23           & No difference    \\ \hline 
        $\Delta$TPR & 1 & active             & 0.024 ± 0.003      & 0.022 ± 0.002     & 0.48             & 0.70            & No difference    \\ \hline  
        $\Delta$TPR & 2 & inactive           & -0.001 ± 0.003     & -0.006 ± 0.003    & 0.24             & 1.17            & No difference    \\ \hline
        $\Delta$TPR & 2 & active             & 0.035 ± 0.004      & 0.028 ± 0.004     & 0.18             & 1.34            & No difference    \\ \hline
        $\Delta$FPR & 1 & inactive           & -0.023 ± 0.004     & -0.021 ± 0.004    & 0.62             & -0.50           & No difference    \\ \hline 
        $\Delta$FPR & 1 & active             & -0.009 ± 0.003     & -0.007 ± 0.003    & 0.79             & -0.27           & No difference    \\ \hline
        $\Delta$FPR & 2 & inactive           & -0.021 ± 0.004     & -0.030 ± 0.005    & 0.18             & 1.34            & No difference    \\ \hline 
        $\Delta$FPR & 2 & active             & -0.015 ± 0.005     & -0.018 ± 0.005    & 0.61             & 0.51            & No difference    \\ \hline
        $\Delta$SHD & 1 & inactive           & -0.033 ± 0.013     & -0.047 ± 0.010    & 0.36             & 0.91            & No difference    \\ \hline 
        $\Delta$SHD & 1 & active             & -0.072 ± 0.011     & -0.056 ± 0.010    & 0.30             & -1.04           & No difference    \\ \hline  
        $\Delta$SHD & 2 & inactive           & -0.082 ± 0.013     & -0.086 ± 0.013    & 0.82             & 0.23            & No difference    \\ \hline 
        $\Delta$SHD & 2 & active             & -0.126 ± 0.016     & -0.100 ± 0.017    & 0.28             & -1.09           & No difference    \\ \hline
    \end{tabular}
    % \end{center}
    \label{tab:full_4_imp_vs_exp}
\end{table*}
% \end{table}
%
Table~\ref{tab:full_4_imp_vs_exp} shows the detailed results of the difference between empirical improvements due to knowledge induction vs expected outcomes using misclassified edge set as the knowledge source.
%
% \begin{table}
\begin{table*}[tb]
    \centering
    \caption{Full results for inducing knowledge in real data (Section~\ref{subsection:experiments_real_data}).} 
    % \begin{center}
    \begin{tabular}{|c|c|c|c|c|c|}
    % \begin{tabular}{@{}rrrrrr@{}}
        \hline
        Metric                           & $\Delta$ & Mean ± Stderr. & p-value & t-stat & Remarks \\ 
        \hline
        $\Delta$TPR & 1          & 0.020 ± 0.004           & 8.10E-06         & 4.60            & Improvement      \\  
        \hline
        $\Delta$TPR & 2          & 0.036 ± 0.005           & 1.77E-12         & 7.62            & Improvement      \\ 
        \hline
        $\Delta$ \%   edge in agreement & 1          & 0.393 ± 0.086           & 8.10E-06         & 4.60            & Improvement      \\ 
        \hline
        $\Delta$ \%   edge in agreement & 2          & 0.714 ± 0.094           & 1.77E-12         & 7.62            & Improvement      \\ \hline
        $\Delta$ \%   edge reversed & 1          & -0.073 ± 0.030          & 1.54E-02         & -2.45           & Improvement      \\ 
        \hline
        $\Delta$ \%   edge reversed & 2          & -0.107 ± 0.033          & 1.29E-03         & -3.27           & Improvement      \\ 
        \hline
    \end{tabular}
    % \end{center}
    \label{tab:full_4_real_data}
\end{table*}
% \end{table}
%
Table~\ref{tab:full_4_real_data} shows the detailed results for inducing knowledge on the real dataset (from \cite{sachs2005causal}).

\end{document}